\pdfoutput=1
\documentclass[11pt]{article}

\usepackage[final]{acl}
\usepackage{times}
\usepackage{latexsym}

\usepackage[T1]{fontenc}
\usepackage[utf8]{inputenc}

\usepackage{microtype}

\usepackage{inconsolata}

\usepackage{graphicx}

\usepackage{color}
\usepackage{color,soul}
\usepackage{xcolor}
\usepackage{comment}
\usepackage{amssymb}
\usepackage{bbding}

\usepackage{multirow}
\usepackage{booktabs}
\usepackage{caption}

\usepackage{enumitem}
\usepackage{makecell}
\usepackage{longtable}
\usepackage{xtab}
\usepackage{titlesec}
\usepackage{tabularx}
\usepackage{hyperref}

\usepackage[most]{tcolorbox}
\newcommand{\fon}[1]{\fontfamily{#1}\selectfont}
\usepackage{colortbl}
\usepackage{graphicx}
\lstdefinelanguage{prompt}{
  keywords={ SHOTS, CURRENT_STATES, CURRENT_STATE_DESCRIPTION, ACTION_DESCRIPTION, PRECONDITION, CORESTATES, THOUGHT, STATE_TO_TRANSFER},
  keywordstyle=\color{blue}\bfseries,
  sensitive=true,
  breaklines=true,
  columns=fullflexible,
  basewidth = {.6em},
  breakindent = {0em},
  tabsize=1,
  aboveskip=0em,
  belowskip=0em,
  comment=[l]{>},
  commentstyle=\color{purple}\ttfamily,
  stringstyle=\color{blue}\ttfamily
}

\newcommand{\Nan}[1]{\textcolor{black}{#1}}

\title{BeSimulator: A Large Language Model Powered Text-based Behavior Simulator}

\author{
    Jianan Wang$^{1,2}$, 
    Bin Li$^{2*}$, 
    Jingtao Qi$^{2}$, 
    Xueying Wang$^{2}$, 
    Fu Li$^{2}$,     
    Hanxun Li$^{1,2}$ \\
    $^{1}$College of Computer Science and Technology, National University of Defense Technology \\
    $^{2}$ Intelligent Game and Decision Lab (IGDL) \\
  \texttt{wangjianan@nudt.edu.cn} ~~~~~ \texttt{libin\_bill@126.com}}

\begin{document}
\maketitle

{\renewcommand{\thefootnote}{}
\footnotetext{*Corresponding authors}}

\begin{abstract}
% background
Traditional robot simulators focus on physical process modeling and realistic rendering, often suffering from high computational costs, inefficiencies, and limited adaptability. 
% behavior simulation
% To handle this issue, we propose Behavior Simulation in robotics to emphasize checking the behavior logic of robots and achieving sufficient alignment between the outcome of robot actions and real scenarios.
To handle this issue, we concentrate on behavior simulation in robotics to analyze and validate the logic behind robot behaviors, \Nan{aiming to achieve preliminary evaluation before deploying resource-intensive simulators and thus enhance simulation efficiency.}
%sufficient alignment between the outcomes of robot actions and real-world scenarios.
% our work
In this paper, we propose \textbf{BeSimulator}, a modular and novel LLM-powered framework, as an attempt towards behavior simulation in the context of text-based environments. 
By constructing text-based virtual environments and performing semantic-level simulation, BeSimulator can generalize across scenarios and achieve long-horizon complex simulation. 
\Nan{Inspired by human cognition paradigm, it employs a ``consider-decide-capture-transfer'' four-phase simulation process, termed \textit{Chain of Behavior Simulation (CBS)}, which excels at analyzing action feasibility and state transition.} 
\Nan{Additionally, BeSimulator incorporates code-driven reasoning to enable arithmetic operations and enhance reliability, and reflective feedback to refine simulation.}
% results 
Based on our manually constructed behavior-tree-based simulation benchmark, BTSIMBENCH, our experiments show a significant performance improvement in behavior simulation compared to baselines, ranging from 13.60\% to 24.80\%.
Code and data are available at \href{https://github.com/Dawn888888/BeSimulator}{https://github.com/Dawn888888/BeSimulator}.
\end{abstract}

\section{Introduction}

% figure 1
\begin{figure}[htbp]
    \centering
    \includegraphics[width=1.0\linewidth]{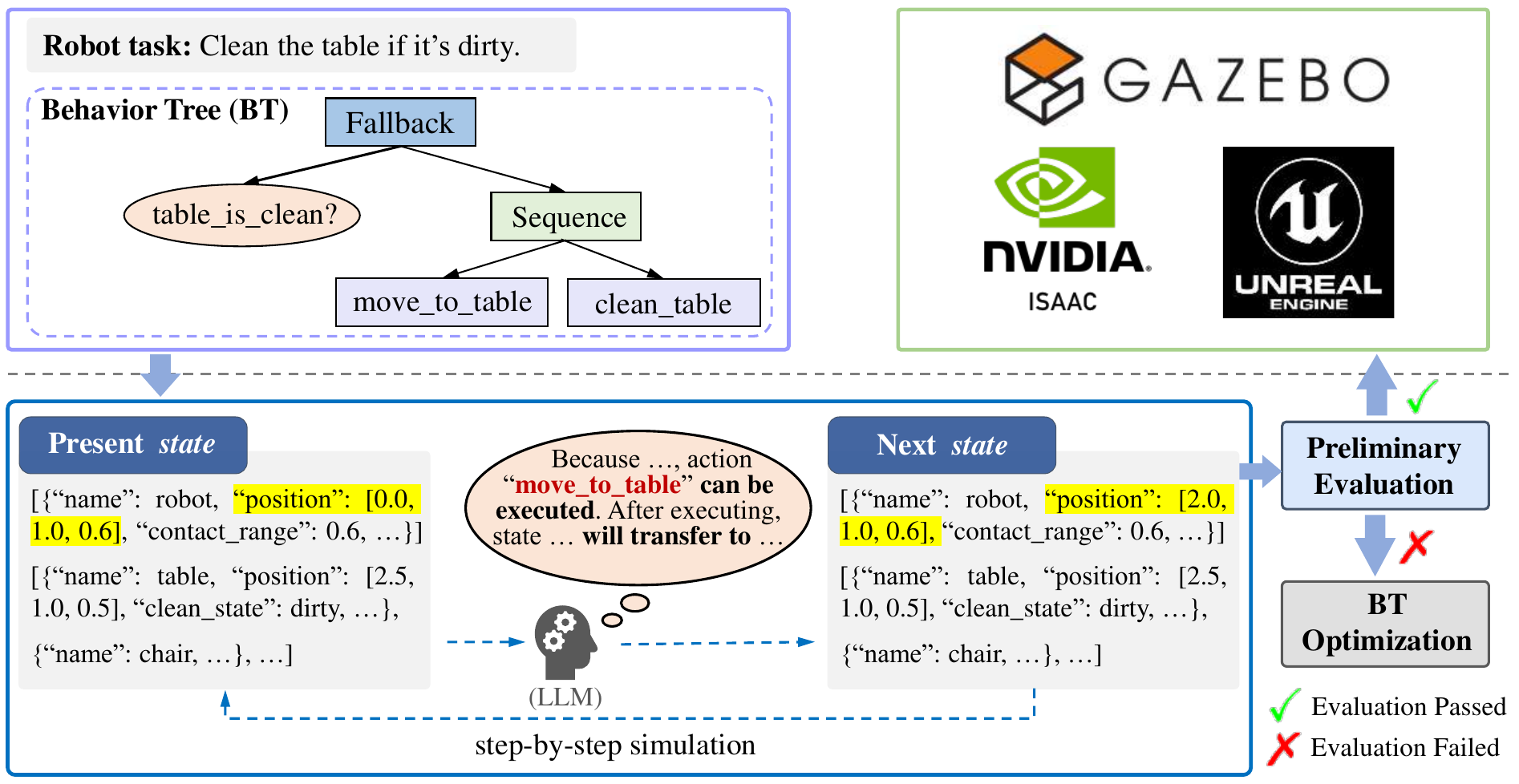}
    \caption{
    Workflow of BeSimulator.
    Based on the task description and robot behavior planning (e.g., BTs), BeSimulator employs LLMs to conduct text-based simulations for identifying behavior logic defects. This serves as a preliminary evaluation before using conventional simulators, enhancing simulation efficiency.}
    \label{fig:figure1}
\end{figure}

%Simulation; Current simulator; Challenge
Simulation plays a pivotal role in robotics, providing a controlled platform for testing, enabling researchers to iteratively optimize robotic systems while circumventing the risks and costs associated with physical prototyping \cite{koenig2004design}. 
%Simulation plays a pivotal role in robotics by providing a controlled platform for iterative testing and optimization of robotic systems, thus circumventing the inherent risks and expenses associated with physical prototyping \cite{koenig2004design}. 
Conventional simulation tools, such as Gazebo \cite{Gazebo} and Unreal Engine \cite{unreal-engine}, have been extensively utilized for tasks involving navigation and human-robot interaction \cite{takaya2016simulation, chandan2021arroch}. 
However, these platforms are predominantly domain specific and focus on modeling physical processes as well as pursuing realistic rendering, which struggle with high computational demands, inefficiencies, and restricted adaptability to the dynamic and diverse nature of environments \cite{staranowicz2011survey}. 
Additionally, they rely on domain experts to design initial simulation scenes and evaluate the results.

%behavior simulation
\Nan{To alleviate this problem, we focus on behavior-level simulation, which abstracts complex physical interactions between robots and environments while maintaining action outcomes consistent with real-world scenarios. 
It emphasizes the control logic of robotic behavior planning solutions (BPS), such as Finite State Machine (FSM) \cite{lee1996principles}, Hierarchical Task Network (HTN) \cite{hayes2016autonomously}, and Behavior Tree (BT) \cite{colledanchise2018behavior}, aiming to detect the potential conflicts with reality and task logic.
Compared to conventional simulators, behavior-level simulation offers greater computational efficiency and broader generalizability, enabling early detection of behavior logic defects before deploying conventional simulators, thus reducing robotic development time and cost.}

%To alleviate this problem, we concentrate on behavior-level simulation, which supports preliminary validation and evaluations before deploying resource-intensive simulators, as shown in Figure \ref{fig:figure1}.
%Behavior-level simulation concentrates on the control logic of behavior planning solutions (BPS) of robots, like Finite State Machine (FSM) \cite{lee1996principles}, Hierarchical Task Network (HTN) \cite{hayes2016autonomously}, and Behavior Tree (BT) \cite{colledanchise2018behavior}. 
%By pursuing action outcomes that are sufficiently similar to real scenarios, behavior-level simulation can efficiently simulate BPS, check whether their control logic matches reality and task logic, and achieve generalization across behavior planning scenarios.
%\Nan{Thereby, it helps to reduce development costs and expedite the robotics development cycle.}

%LLMs as World mode; text-based; LLMs for simulation
\Nan{World model, which originates from the mental models of humans, can predict the next \textit{state} after executing an \textit{action} in the present state \cite{ha2018recurrent, matsuo2022deep}.
Trained on large-scale datasets, Large Language Models (LLMs) encapsulate rich world knowledge and have exhibited potential as sophisticated world models for reasoning and planning \cite{hao2023reasoning, zhao2024large}.
This raises a pivotal research question: can the intrinsic world modeling capacity of LLMs be utilized for robotic behavior simulation?}
\Nan{Existing research \cite{wang2024can} has demonstrated its feasibility in text-based environments.
However, experiments show that while the direct simulation performance of LLMs is impressive, their reliability remains limited. 
The limitation mainly stems from two factors: the inability to capture state transitions that are indirectly related to actions and challenges in arithmetic reasoning.}

%Existing research \cite{wang2024can} has demonstrated its feasibility in text-based context.
%Its experiment results indicate that the direct simulation performance of LLMs is impressive but unreliable, which is due to the inability to capture state transitions indirectly related to actions, as well as transitions that need arithmetic or scientific reasoning.

%our work
To bridge this gap, we propose \textbf{BeSimulator}, a LLM-powered framework designed to efficiently simulate BPS, as an effort towards behavior simulation in text-based environments, as shown in Figure \ref{fig:figure1}. 
BeSimulator consists of three key modules:
%-
1) \textit{Case Generation}, to generate the text-based simulation environment from the robot task, which includes diverse world states;
%--
2) \textit{BPS Simulation}, to perform step-by-step behavior simulation according to the control logic of the solution, which means conducting state transitions on the generated case states;
%---
3) \textit{BPS Evaluation}, to evaluate the effectiveness of the solution and think about potential defects if ineffective.
In contrast to conventional simulation tools, it integrates environment design and result evaluation, thus alleviating the costly need for expert involvement.
To improve the LLMs reliability, BeSimulator adopts \textit{Chain of Behavior Simulation (CBS)}—a four-phase simulation process inspired by human cognitive reasoning paradigm—to deeply analyze the action feasibility and state transitions, especially those indirectly associated with actions.  
It also incorporates a code-driven reasoning mechanism to tackle arithmetic reasoning challenges. 
Moreover, a reflective feedback mechanism is utilized to enhance the LLMs' error recovery capability, thus refining simulation.

%benchmark
Given the modularity and popularity of BTs in robotic control, we construct a BT simulation benchmark, BTSIMBENCH, to evaluate our approach's performance. 
This benchmark provides 75 BTs based on the 25 robot tasks from BEHAVIOR-1K\cite{li2024behavior}, which are long-horizon and rely on complex manipulation skills. 
Experimental results indicate that BeSimulator enhances the reliability of LLMs in behavior simulation, achieving higher accuracy in identifying defective behavior planning compared to the baselines.

% contributions
Overall, we make the following contributions:
\setlist[itemize]{itemsep=-3pt, topsep=0pt}
\begin{itemize}
    
    \item We propose BeSimulator, an LLM-powered, text-based behavior simulator to efficiently analyze and validate the behavior logic of robots. Inspired by human cognition processes, it integrates the Chain of Behavior Simulation to enhance analyzing action feasibility and state transitions.

    \item Considering inherent biases and struggle with numeric reasoning of LLMs, BeSimulator incorporates the code-driven reasoning and reflective feedback mechanisms to enhance reliability and refine simulation.

    \item We construct BTSIMBENCH, a simulation benchmark based on BTs, containing various task categories and behavior types. Experimental results on BTSIMBENCH across four LLMs demonstrate BeSimulator's versatility and effectiveness in behavior simulation.
    
\end{itemize} 
\section{Related Works}

\subsection{Robotics Simulators}
%Compared to validating on real robots, robot simulators provide a secure platform for the evaluation of new strategies, solutions, or algorithms, which accelerate robot development and save costs\cite{koenig2004design}. 
%Currently, common robot simulators include general simulators like Gazebo\cite{Gazebo}, Isaac Sim\cite{Isaac-Sim}, V-REP\cite{rohmer2013v}, and some game engines like Unreal Engine\cite{unreal-engine}.
\Nan{Common robot simulators include general simulators like Gazebo \cite{Gazebo}, Isaac Sim \cite{Isaac-Sim}, V-REP \cite{rohmer2013v}, and some game engines like Unreal Engine \cite{unreal-engine}. }
Robot simulators have been widely used for tasks related to navigation \cite{takaya2016simulation}, human-robot interaction \cite{chandan2021arroch}, vehicle driving \cite{dosovitskiy2017carla}, etc. 
Their performance is primarily determined by the physics and rendering engine. 
The physics engine is used to mathematically model complex physical processes like motion, collisions, etc. 
The rendering engine provides a visual interface to enhance simulation realism.
%However, these simulators are limited to low efficiency, high computational cost, poor generalization \cite{staranowicz2011survey, iovino2021learning}, and dependency on manual scene construction and expert evaluation, which our study aims to tackle in text-based environment.
\Nan{However, these simulators suffer from low efficiency, high computational costs, poor generalization capability \cite{staranowicz2011survey, iovino2021learning}, and reliance on manual scene construction and expert evaluation. }
\Nan{Based on this, we focus on behavior-level simulation to examine logical defects behind robotic behaviors, thereby achieving preliminary validation before deploying resource-intensive simulators. 
This can significantly reduce cycles and associated costs in robotic system development.}
%which our study aims to solve.

\subsection{Conventional Behavior Simulation}
Traditionally, behavioral simulation refers to simulating the behavior patterns of specific subjects. 
Most researches focus on user-behavior simulation to evaluate the effectiveness of evacuation systems \cite{kountouriotis2016icrowd, harada2015switching}. 
Moreover, some researchers focus on behavior simulation of various subjects, like infants \cite{nishida2004infant}, humans living in atypical buildings \cite{lee2019actoviz}, etc. 
There are also research efforts \cite{zhang2023user, hassouni2018using} that utilize behavior simulators to provide a simulation environment for reinforcement learning (RL). These are fundamentally different from the behavior simulation expounded in this paper.

\subsection{LLMs for Logical Reasoning}
Trained on the large-scale corpus, LLMs exhibit remarkable reasoning capability. 
LLMs are typically prompted to decompose complex problems and engage in step-by-step thinking and reasoning, exemplified by CoT \cite{wei2022chain}, Zero-shot-CoT \cite{kojima2022large}, self-consistency \cite{wang2022self}, etc. 
Some methods combine reasoning problems with search algorithms like ToT \cite{yao2024tree}, RAP \cite{hao2023reasoning}, etc. 
Moreover, some researchers focus on supervised fine-tuning of LLMs to improve reasoning, such as WOMD-Reasoning \cite{li2024womd}, CPO \cite{zhang2024chain}, etc. 
Methods like CoC \cite{li2023chain} and ToRA \cite{gou2023tora} apply external tools such as code interpreters, computation libraries, etc., aiming to reduce computational hallucination and enhance reasoning. 
\Nan{Reflect \cite{liu2023reflect} utilizes generated failure explanations to rectify reasoning and planning errors.}
In this paper, we propose Chain of Behavior Simulation containing four phases from shallow to deep, which conforms to the human cognition process further.
\section{Problem Formalization}

Based on research \cite{colledanchise2018behavior, cai2021bt}, a behavior planning problem can be formalized as a quintet: \textless $\mathcal{S}$, $\mathcal{A}$, $T$, $s_{initial}$, $g$\textgreater, where $\mathcal{S}$ is state space, $\mathcal{A}$ is action space, $T: \mathcal{S} \times \mathcal{A} \rightarrow \mathcal{S}$ is the state transition rules, $s_{initial}$ is the initial scene state and $g$ is goal condition. 
After executing action $a$ in the state $s_{t}$, the next state $s_{t+1}$ = $T(s_{t}, a)$. 
The target of the behavior planning problem is to produce a solution $\pi$ capable of transferring $s_{initial}$ to $g$ in finite steps. 
In this paper, we aim to determine whether one behavior planning solution is effective through behavior simulation. 
Our main idea is that based on the initial scene states $s_{initial}$ and the control logic of the solution $\pi$, step by step perform state transitions and finally determine whether the goal condition $g$ is achieved, as is shown in Eq \ref{next_state}. 
In this paper, we integrate LLMs to implement this process to achieve automated and long-horizon behavior simulation.
\begin{equation}
    \label{next_state}
    g \subseteq T(s_{initial}, \pi)
\end{equation}
\section{Method}

% A.Overview
\subsection{Overview}
\begin{figure*}
    \centering
    \includegraphics[width=\linewidth]{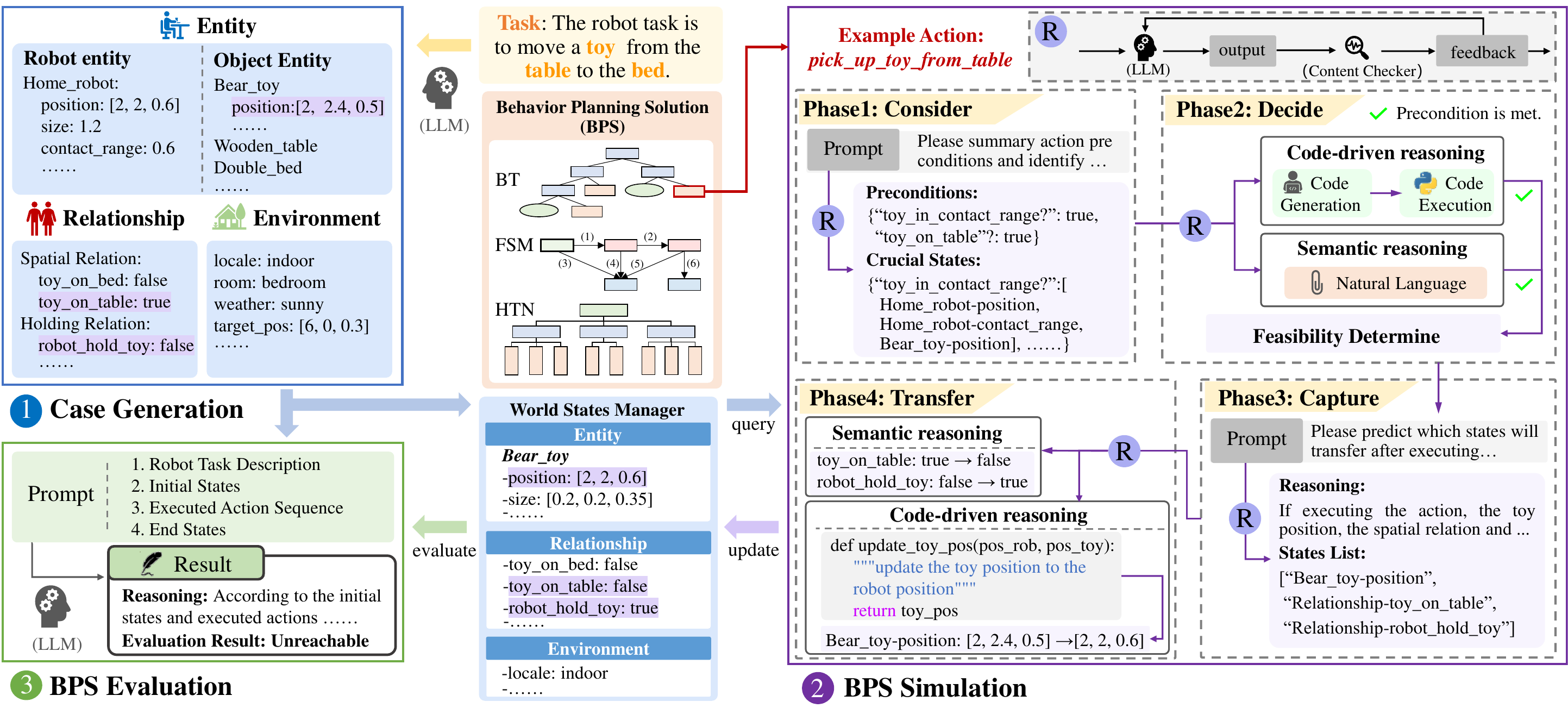}
    \caption{Overview of BeSimulator. 
    \textbf{Module 1: Case Generation} generates simulation cases from task descriptions. The cases contain diverse world states, which are subsequently maintained by a world state manager.
    \textbf{Module 2: BPS Simulation} dynamically simulates action sequences according to the execution logic of BPS. Using single action \textit{pick\_up\_toy\_from\_table} as an example, the schematic illustrates the four-phase ``consider-decide-capture-transfer'' process, which first checks the action feasibility and performs state transitions to update the manager. BeSimulator integrates code-driven reasoning and reflective feedback into the process.
    \textbf{Module 3: BPS Evaluation} conducts evaluation based on the simulation results.}
    \label{fig:overview}
\end{figure*}

In this paper, we propose BeSimulator, illustrated in Figure \ref{fig:overview}, a modular and novel LLM-powered framework that conducts behavior simulation in text-based environments. 
%Through generating scene cases and implementing semantic-level simulation, BeSimulator is general across different domains. 
It consists of three key modules, including \textit{Case Generation}, \textit{BPS Simulation}, and \textit{BPS Evaluation}. 
Based on robot tasks, BeSimulator first generates simulation cases, including various world states. 
Then, according to the control logic of the BPS, BeSimulator analyzes the action feasibility and performs state transitions step by step.
\Nan{Finally, in view of the simulation results, BeSimulator evaluates whether the solution is effective.}

% Figure: overview

% B.Case Generation
\subsection{Case Generation}
Given a robot task, \Nan{BeSimulator first initiates the corresponding simulation case.
Leveraging the scene comprehension and generation capabilities of LLMs, the case generation exhibits strong generalization and is conducive to constructing complex scenes.
The three parts of a case are as follows.}

\begin{itemize}
    \item \textbf{Entity Information} is composed of robot entities and object entities. 
    \Nan{The robot entities are autonomous and capable of action, e.g., \textit{sweeping robots}, \textit{quadruped robots}.}
    The object entities are static, e.g., \textit{tables}, \textit{chairs}. 
    \Nan{For each entity, the generated states include basic properties: id, type, position, size, and task-related properties.
    E.g., if the task requires the robot to turn on lamp, the \textit{on\_off\_state} property of the lamp is considered task-related.}
    %Note that these states are descriptive, enabling LLMs to directly infer their definitions literally.
    
    \item \textbf{Relationship Information} \Nan{defines interactions between entities, including spatial relationships (e.g., \textit{on}, \textit{in}, \textit{distance}), holding relationships, manipulable relationships, etc.}
        
    \item \textbf{Environment Information} \Nan{describes task-specific settings, e.g., \textit{locale}, \textit{weather}, \textit{object target poses}.}  
\end{itemize}

% C.BPS Simulation
\subsection{BPS Simulation}
\Nan{Building upon the initial scene case, BeSimulator dynamically simulates action sequences based on the control logic of the BPS.} 
Our approach leverages the reasoning capability and embedded knowledge of LLMs to support behavior-level simulation while significantly improving efficiency. 
\Nan{To address the reliability limitations of direct LLM-based simulation, we propose CBS, an atomic action simulation process inspired by the human cognition processes. 
Additionally, we incorporate code-driven reasoning to ensure precise numerical computation and employ reflective feedback mechanism to iteratively refine simulation.}

% CBS
\subsubsection{Chain of Behavior Simulation}
For each action, \Nan{BeSimulator employs chain of behavior simulation to model its execution via a structured four-phase process: consider-decide-capture-transfer.
Through CBS, BeSimulator first analyzes the action's feasibility (phase 1-2) and, if feasible, predicts the resulting state transitions (phase 3-4).}
The prompt designs for CBS are detailed in Appendix \ref{appendix:PROMPT}.

During the first two phases, \Nan{BeSimulator analyzes whether the action execution adheres to the real-world logic.} 
Specifically, in the \textbf{consider} phase, \Nan{BeSimulator ponders the preconditions required for successful action execution} and identifies the crucial states affecting each precondition. 
Subsequently, in the \textbf{decide} phase, \Nan{BeSimulator assesses the satisfaction of each precondition by examining the corresponding crucial states.} 
Then it aggregates these assessment results to determine the action's feasibility within the current scene.
For instance, when simulating the action \textit{``pick\_up\_toy''}, \Nan{BeSimulator summarizes two preconditions}: \textit{``can\_robot\_touch\_toy?=true''} and \textit{``whether\_robot\_has\_free\_gripper?=true''}. 
For the first precondition, \Nan{the crucial states inferred include} \textit{robot\--position}, \textit{toy-position} and \textit{robot-gripper\_contact\_range}, \Nan{which determine whether the precondition is satisfied}. 

\Nan{Upon confirming an action's executability, BeSimulator predicts the state transitions through the last two phases.} 
Detailedly, in the \textbf{capture} phase, \Nan{it identifies which scene states will be affected by the action}.
\Nan{Then, BeSimulator determines the precise transition rules for each impacted state in the \textbf{transfer} phase.} 
For example, for action \textit{``move\_to\_bed''}, the states to be updated may include \textit{robot-position}, \Nan{the positions of held objects, the involved spatial relationships — with the latter two are indirectly related to the action.}
Moreover, BeSimulator utilizes a world states manager to implement state transitions. 
\Nan{This manager, which maintains all scene states in a structured representation and supports the state querying and updating functionalities}, serves as a textual virtual environment and effectively \Nan{reduces LLMs' hallucinations regarding the diverse scene states.}

Compared to behavior simulation with just one phase, \Nan{our four-phase reasoning paradigm decomposes the complex simulation problem through sequential, human-like cognitive processes.}
\Nan{This significantly improves the analysis of action feasibility and state transitions (especially indirectly induced transitions), and enhances the LLMs simulation reliability, which is the core goal of our approach.}
We further investigate its effectiveness through ablation study in Section \ref{sec:ablation_cbs}.

% Code-driven Reasoning
\subsubsection{Code-driven Reasoning}
While LLMs demonstrate strong capabilities in semantic reasoning, their performance often degrades when handling tasks requiring numerical reasoning. 
To address this limitation, \Nan{BeSimulator employs a code-driven reasoning mechanism for arithmetic operations involving numerical states. }
\Nan{This mechanism, applied during the decide and transfer phases, integrates code generation and code execution via a code interpreter to ensure computational precision and improve simulation fidelity.}
\Nan{Taking decide phase as an example, BeSimulator dynamically selects its reasoning mode based on data types of critical states.
If these states involve numerical types, it activates code-driven reasoning; otherwise, it defaults to semantic reasoning.}
Ablation experiment in Section \ref{sec:ablation_cg} validates its efficacy.

% Reflective Feedback
\subsubsection{Reflective Feedback}
\Nan{Required to respond in JSON format, LLMs occasionally produce errors in syntactic accuracy and semantic consistency.}
Therefore, BeSimulator incorporates a reflective feedback mechanism.
After LLMs generate an output, \Nan{an automated content checker evaluates its validity from two aspects.}
\Nan{First, the checker performs syntactic validation by examining four key aspects: (1) adherence to JSON format, (2) completeness of JSON keys, (3) accuracy of JSON values, and (4) executability of the generated codes.}
\Nan{Second, the checker conducts semantic validation by assessing whether the reasoning process aligns logically with the final output, identifying potential inconsistencies.}
If errors are detected, BeSimulator provides feedback to the LLM, guiding it to reflect and re-output.
\Nan{The process iterates until either all errors are resolved or a predefined feedback limit is reached.}
In Section \ref{sec:ablation_rf}, we demonstrate the effectiveness of the mechanism through ablation.

\begin{table}[htbp]
    %\captionsetup{skip=5pt}
    \centering
    \small
    \begin{tabular}{l c c}
        \toprule
        \textbf{Category}    & \textbf{Reality logic} & \textbf{Task logic} \\
        \midrule
        Good              & \textcolor{green}{\checkmark} & \textcolor{green}{\checkmark} \\
        Counterfactuals   & \textcolor{red}{\XSolidBrush} & N/A \\
        Unreachable       & \textcolor{green}{\checkmark} & \textcolor{red}{\XSolidBrush}   \\   
        \bottomrule
    \end{tabular}
    \caption{\label{tab:category}
    The three categories of the evaluation results. \textcolor{green}{\checkmark} Consistent with logic. \textcolor{red}{\XSolidBrush} Inconsistent with logic.}
\end{table}

% D.BPS Evaluation
\subsection{BPS Evaluation}

\Nan{Based on the simulation results, hSimulator classifies BPS into three distinct categories: Good, Counterfactuals, and Unreachable, as presented in Table \ref{tab:category}. 
The latter two categories indicate inherent logic defects in the solution.}

Specifically, during the step-by-step simulation process, if any action is infeasible—indicating conflicts between the solution's execution logic and reality logic—the solution is classified as Counterfactuals. 
\Nan{Moreover, for solutions that maintain logical consistency with reality, BeSimulator performs rigorous evaluation based on four key elements: task objectives, initial scene states, executed action sequences, and terminal states.} 
The evaluation yields either Good or Unreachable classification.
\Nan{The Good category indicates that the solution achieves the task goal at the behavior level, while Unreachable indicates fundamental incompatibility between the solution and task requirements. }
\Nan{Consider the example of \textit{``Clean book with rag''} task, which presents two defective solutions, namely Solution A and Solution B.
Solution A, which controls the robot to perform the action \textit{``move\_to\_book''} followed by \textit{``clean\_book''}, exhibits Counterfactuals because it omits the crucial step of picking up the rag before cleaning.
In contrast, Solution B is classified as Unreachable because the robot only picks up the rag and the book after executing the complete solution, thereby failing to achieve the intended task goal.}

\section{Experiments}
\label{sec:experiments}

% overview
\Nan{Our approach is versatile and can be adapted across different BPS. 
In this section, we use behavior tree (BT) as a case study and perform experiments to evaluate the effectiveness of our approach. 
We first propose a novel BT simulation benchmark and then conduct comprehensive experiments to address two research questions:} 
\begin{itemize}
    %\item Can BeSimulator improve LLMs' performance in behavior simulation? (Section \ref{sec:performance}) 
    \item To what degree of accuracy does BeSimulator conduct behavior simulation? (Section \ref{sec:performance}) 
    \item To what extent do our designed mechanisms contribute to the simulation, including chain of behavior simulation, code-driven reasoning, and reflective feedback? (Section \ref{sec:ablation})
\end{itemize}

% Tab: BTs nodes
%\vspace{-3pt}
%\renewcommand{\arraystretch}{1.1}
\begin{table}[htbp]
{
    %\captionsetup{skip=5pt}
    \centering
    \small
    \renewcommand\tabcolsep{7pt}
    \begin{tabularx}{\linewidth}{p{0.15\linewidth} p{0.5\linewidth} p{0.15\linewidth}}
    
        \toprule
        \textbf{Node}  & \textbf{Descriptions} & \textbf{Execution}                    \\
        \midrule
        
        \multirow{2}{*}{Sequence}  &  ticks its child nodes from left to right until one returns \textit{Failure} & \multirow{6}{*}{\parbox{0.15\linewidth}{py\_trees rules}} \\ [0.7ex]
        \cline{2-2}
        \addlinespace[0.7ex]
        
        \multirow{2}{*}{Fallback}  &  ticks its child nodes from left to right until one returns \textit{Success} &        \\ [0.7ex]
        
        \cline{2-2}
        \addlinespace[0.7ex]
        \multirow{2}{*}{Parallel}  &  ticks its child nodes in parallel and returns based on the setting          &        \\
        
        \midrule       Action    &  performs an action            & CBS         \\
        Condition &  checks if a condition is met      & CBS         \\
        \bottomrule
        
    \end{tabularx}
    \caption{\label{tab:bt}
    Typical nodes in BTs and the execution rules they follow.}
}
\end{table}

% A.BTs Simulation
%\vspace{-7pt}
\subsection{BTs Simulation}
A BT is a directed tree structure where leaf nodes (Condition and Action nodes) control the robot's perception and actions, while internal nodes (e.g., Fallback and Sequence nodes) manage the execution logic of leaf nodes \cite{colledanchise2018behavior}, as is shown in Table \ref{tab:bt}.
BT execution begins at the root node, which ticks its descendant nodes at each time step through Depth First Search (DFS).  
Based on the current scene, the tick creates a control flow that determines the robot's perception and action sequence.
  
In our experiments, we utilize py\_trees\footnote{\url{https://py-trees.readthedocs.io/en/devel/}} as BTs engineer that sends tick signal and controls nodes execution. 
Internal nodes follow the execution rules defined in py\_trees, while leaf nodes are handled by the the CBS mechanism. 
\Nan{Specifically, BeSimulator implements the four-phase CBS process for action node simulation and employs a two-phase variant for condition nodes.
This variant adopts the first two CBS phases, ``consider-decide'': it first identifies relevant scene states for the condition node, then determines its output (Success/Failure) based on their values.
For example, for the condition node ``is\_near\_book?'', BeSimulator infers critical states (e.g., \textit{robot-position}, \textit{book-position}, and \textit{robot-contact\_range}) and generates code to determine whether the node succeeds or fails.}

\setlength{\cmidrulewidth}{0.8pt}
\begin{table*}[htbp]
    \centering
    \small
    \setlength{\tabcolsep}{2.2mm}   
    
    \begin{tabular*}{\textwidth}{l cc cccc}
        \toprule
            \multirow{2}{*}{\textbf{Model}}                 & \multirow{2}{*}{\textbf{Method}} & \multirow{2}{*}{\textbf{Delivery(\%)}} & \multicolumn{4}{c}{\textbf{Accuracy(\%)}}         \\
            \cmidrule(lr){4-7}
                           &     &   & \textbf{Good} & \textbf{CFactuals} & \textbf{Unreachable} & \textbf{Average} \\
                                                   
            \midrule
            \multirow{3}{*}{\textbf{Claude-3.5-Sonnet}}     
                & CoT      & 100.00   & $\mathbf{98.40} \pm \mathbf{2.19}$   & $47.20 \pm 4.38$   & $60.00 \pm 2.83$   & 68.53  \\
                & BoN & 100.00  & $\mathbf{98.40} \pm \mathbf{2.19}$   & $50.40 \pm 3.58$   & $64.00 \pm 2.83$   & 70.93  \\
                %& CoT-SC    & 100.00  & $\mathbf{98.40} \pm \mathbf{2.19}$   & $50.40 \pm 2.19$   & $57.60 \pm 2.19$   & 68.80  \\
                & BeSimulator          & 100.00   & $88.80 \pm 3.34$    & $\mathbf{97.60} \pm \mathbf{2.19}$     
                                & $\mathbf{92.00} \pm \mathbf{2.83}$    & $\mathbf{92.80}$    \\
            
            \midrule
            \multirow{3}{*}{\textbf{DeepSeek-V3}}           
                & CoT      & 100.00   & $87.20 \pm 1.79$   & $53.60 \pm 4.56$   & $56.00 \pm 4.00$   & 65.60  \\
                & BoN & 100.00  & $\mathbf{88.80} \pm \mathbf{3.35}$    & $52.00 \pm 2.83$    & $59.20 \pm 3.35$    & 66.67  \\
                %& CoT-SC    & 100.00  & $88.00 \pm 0.00$    & $52.80 \pm 1.79$    & $56.80 \pm 3.35$    & 65.87  \\
                & BeSimulator          & 100.00   & $85.60 \pm 2.19$ & $\mathbf{96.80} \pm \mathbf{3.35}$ 
                                & $\mathbf{88.80} \pm \mathbf{1.79}$               &  $\mathbf{90.40}$   \\
            
            \midrule
            \multirow{3}{*}{\textbf{Qwen2-72B-Instruct}}    
                & CoT      & 100.00   & $92.00 \pm 2.83$   & $2.40 \pm 2.19$   & $52.80 \pm 4.38$   & 49.07  \\
                & BoN & 100.00  & $\mathbf{95.20} \pm \mathbf{1.79}$     & $1.60 \pm 2.19$    & $\mathbf{66.40} \pm \mathbf{2.19}$    & 54.40 \\
                %& CoT-SC    & 100.00  & $\mathbf{96.00} \pm \mathbf{0.00}$       & $1.60 \pm 2.19$     & $64.80 \pm 1.79$        & 54.13 \\
                & BeSimulator          & 100.00   & $68.80 \pm 4.38$ & $\mathbf{90.40} \pm \mathbf{2.19}$ 
                                & $61.60 \pm 3.58$               &  $\mathbf{73.60}$         \\
            
            \midrule
            \multirow{3}{*}{\textbf{Llama3.1-70B-Instruct}} 
                & CoT      & 100.00   & $\mathbf{66.40} \pm \mathbf{4.56}$   & $67.20 \pm 1.79$   & $48.80 \pm 3.35$   & 60.80   \\
                & BoN  & 100.00   & $64.40 \pm 3.58$     & $71.20 \pm 1.79$     & $32.80 \pm 4.38$     & 56.13 \\
                %& CoT-SC     & 100.00   & $65.20 \pm 1.79$     & $71.20 \pm 1.79$     & $24.80 \pm 1.79$    & 53.73 \\
                & BeSimulator          & 100.00   & $65.60 \pm 3.58$ & $\mathbf{89.60} \pm \mathbf{3.58}$ 
                                & $\mathbf{68.00} \pm \mathbf{2.83}$               &  $\mathbf{74.40}$     \\   
        \bottomrule
    \end{tabular*}
    \caption{\label{tab:comparative_experiment}
    The comparative experiment results of BeSimulator on BTSIMBENCH across four LLMs (mean ± standard deviation from five repeated experiments).}
\end{table*}

\subsection{Experimental Setup}

% Figure:BTSIMBENCH
\begin{figure}[t]
    \centering
    \includegraphics[width=0.98\linewidth]{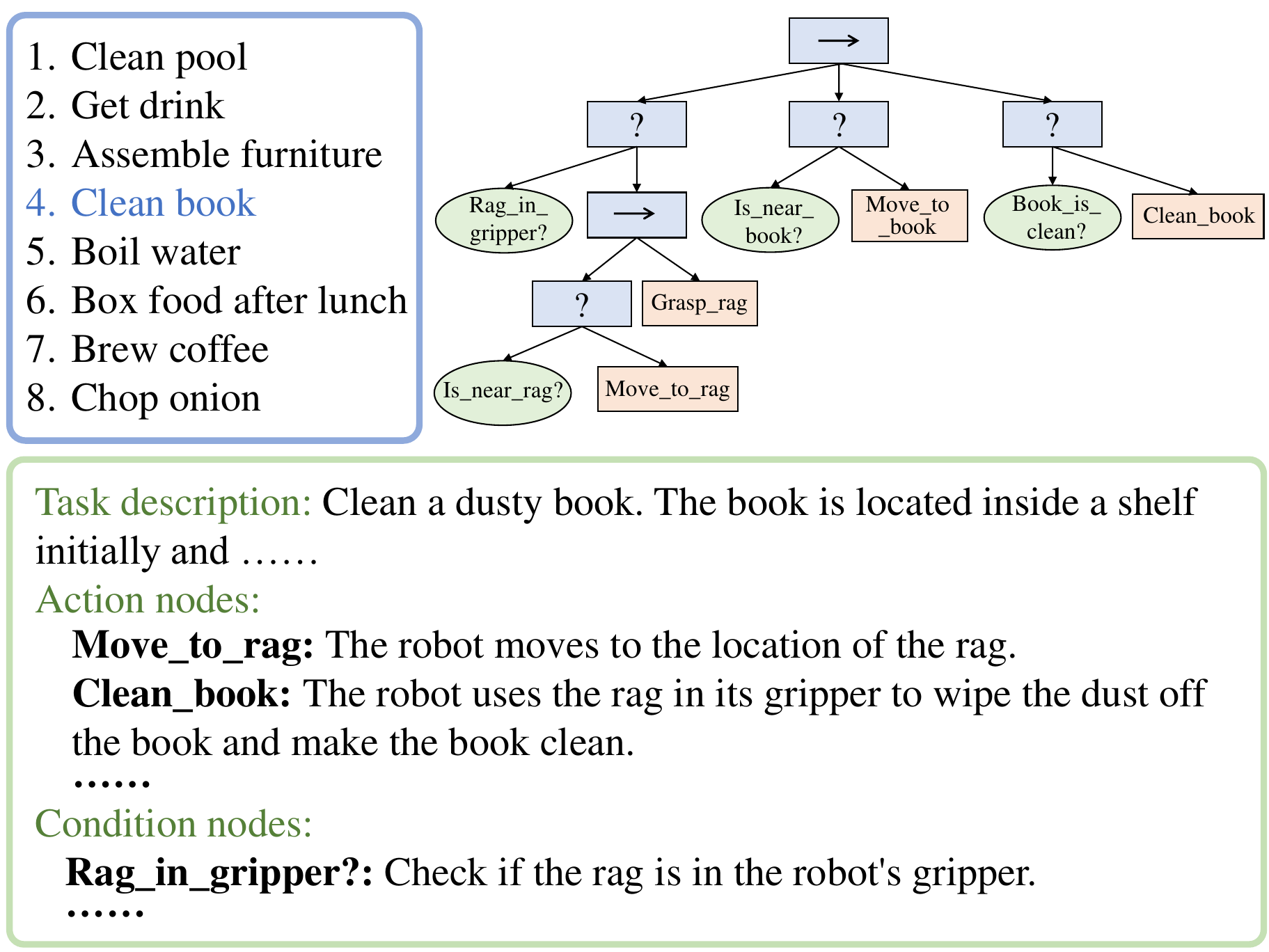}
    \caption{Examples of BTSIMBENCH. 
    Top Left: Example activities. 
    Top right: Good BT for the \textit{Clean book with rag} task, where blue boxes with `$\rightarrow$' and `$?$' denote Sequence and Fallback nodes, respectively. Green ellipses and orange boxes represent Condition and Action nodes. 
    Bottom: Task description and descriptions of Action and Condition nodes in the BT.}
    \label{fig:benchmark}
\end{figure}

% Figure: Bad BTs
\begin{figure}[t]
    \centering
    \includegraphics[width=1\linewidth]{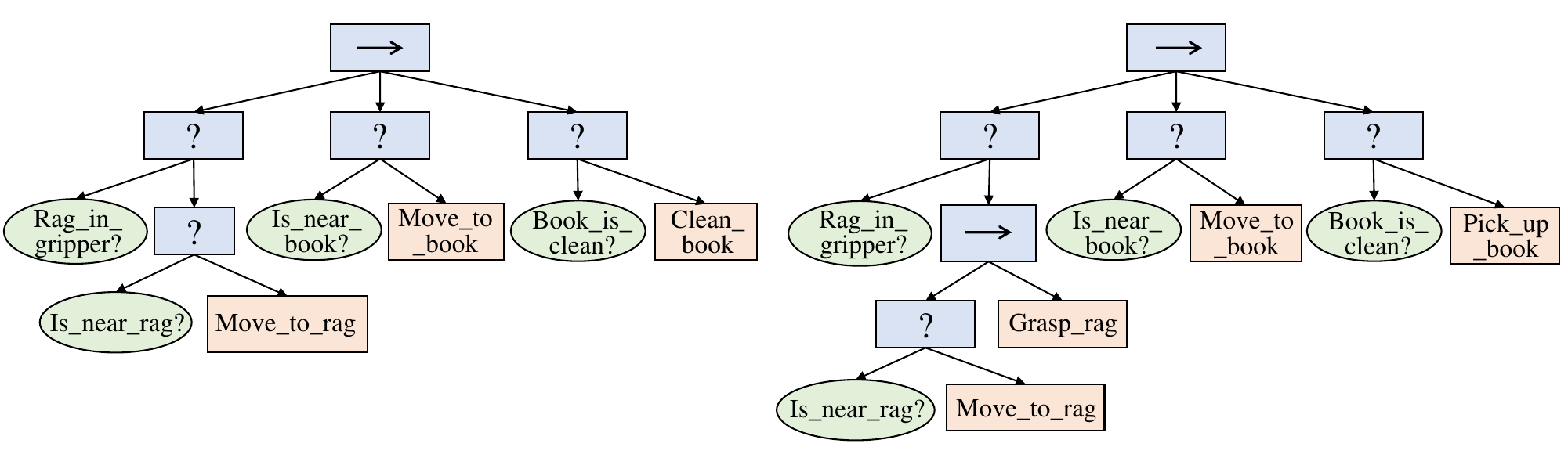}
    \caption{Bad BT examples. Left: BT of the Counterfactuals category. Right: BT of the Unreachable category.}
    \label{fig: bad tree}
\end{figure}

% Benchmarks
\subsubsection{Benchmarks}
Based on BEHAVIOR-1K \cite{li2024behavior} which is a comprehensive benchmark for human-centered robots, we construct BTSIMBENCH, a novel BT simulation benchmark comprising 75 BTs across three categories. 
\Nan{BEHAVIOR-1K includes the definitions of 1000 daily tasks, which are long-horizon and depend on complex manipulation skills. 
These tasks have been experimentally verified to be extremely challenging for current AI algorithms and are widely applied in research on behavior planning \cite{zhang2024dkprompt}, motion control \cite{jiang2025behavior}, etc. 
We select 25 tasks from them while keeping diversity in both task categories and behavior types, then we utilize LLMs to convert them from BEHAVIOR Domain Definition Language (BDDL) into textual descriptions.} 
For each task, we first construct a Good BT and provide the corresponding node descriptions, as illustrated in Figure \ref{fig:benchmark}. 
Additionally, by altering some nodes in the Good BT, we construct a Counterfactuals BT and an Unreachable BT to simultaneously measure the simulation capability of BeSimulator for both good and bad BTs, as shown in Figure \ref{fig: bad tree}.  
For each BT, we conduct multiple rounds of manual verification and adjustment to ensure its availability. 
See Appendix \ref{appendix:BTSIMBENCH} for more BTSIMBENCH info.

% Baseline and Adopted LLMs
\subsubsection{Baselines and Adopted LLMs}
Currently there is no specific framework designed for performing behavior simulation based on LLMs.
Thus, we compare our method with the scenario-agnostic methods including Chain of Thought \cite{wei2022chain} and Best of N with LLM Judge.
Our method and all baselines employ few-shot examples to ensure a fair comparison.

\begin{itemize}
    \item \textbf{Chain of Thought (CoT)}: 
    This method takes the robot task description and the BT with node descriptions as input. Specifically, we instruct LLMs to generate intermediate reasoning steps to analyze the BT's control logic, evaluate its effectiveness, and identify potential reasons if ineffective.

    \item \textbf{Best of N with LLM Judge (BoN)}: 
    According to the CoT method, we sample three candidate answers and then use LLM as a judge to select the best answer based on the correctness of the answers.

    %\item \textbf{CoT-SC}: For this method, we also sample three answers based on the CoT method and select the most consistent answer as the final answer.
\end{itemize}

To assess our method's generalization across LLMs, we select four well-known LLMs in the field of closed source and open source as our base LLMs, including Claude-3.5-Sonnet \cite{Claude-3.5-Sonnet}, DeepSeek-V3 \cite{deepseek-v3}, Qwen2-72B-Instruct \cite{yang2024qwen2} and Llama3.1-70B-Instruct \cite{dubey2024llama}.

% Metrics
\subsubsection{Metrics} 
To evaluate the simulation performance of BeSimulator, we use the following metrics:
\begin{itemize}
    
    \item \textbf{Delivery Rate}: 
    \Nan{This metric assesses whether an LLM-based simulator can successfully deliver a simulation result within finite reflection times of LLMs. Exceeding the predefined feedback limit (5 times in our experiment settings) will result in delivery failure.}
    
    \item \textbf{Accuracy}: 
    This metric represents the proportion of BTs that are correctly evaluated for the corresponding categories. 
    \Nan{As the key evaluation criterion, it reflects the simulator's effectiveness for behavior simulation.}
    
\end{itemize}

% C.Simulation Performance
\subsection{Simulation Performance}
\label{sec:performance}

In our experiments, \Nan{we employ BTSIMBENCH to evaluate the efficacy of BeSimulator across four LLMs. 
Table \ref{tab:comparative_experiment} presents the performance comparison between our method and baselines.}

The results demonstrate that BeSimulator achieves significant performance improvements across all base LLMs, showing its capability for behavior-level simulation.
Specifically, for DeepSeek-V3, we observe a maximum increase of 24.80\% in average accuracy, corresponding to a 37.80\% relative improvement over the CoT baseline. 
Other LLMs exhibit enhancements ranging from 13.60\% to 24.53\%, further validating the efficacy of our approach.
Moreover, the results show that the BoN baseline outperforms CoT in general.
This demonstrates that LLMs are better at making choices based on candidate answers than directly generating answers.

Furthermore, the baselines' strong performance on Good BTs belies a superficial understanding of robot behavior control, as it struggles to detect hidden conflicts with reality and task logic. 
For instance, in the Counterfactuals BT shown in Figure \ref{fig: bad tree}, the baselines often fail to recognize that one precondition for the ``\textit{clean\_book}'' action is ``\textit{hold\_rag?=true}''. 
Consequently, it tends to classify BTs as Good, resulting in degraded performance on faulty BTs.
In contrast, our method outperforms the baselines in the Counterfactuals and Unreachable categories. 
For instance, on the Qwen2-72B-Instruct model, our method achieves 90.40\% accuracy in simulating and evaluating Counterfactuals BTs, while the BoN baseline yields 1.60\% accuracy. 
These results indicate that BeSimulator effectively improves action execution analysis and identifies potential defects, thus facilitating iterative optimization of robot systems.
Additional evaluation results for BeSimulator are provided in Appendix \ref{appendix:Detailed_Exp_Results}.

% Tab: ablation
%\vspace{-5pt}
\begin{table}[tbp]
    \centering
    \fontsize{8.1}{10}\selectfont
    \setlength{\tabcolsep}{1pt}

    \begin{tabular}{l c cccc}
        \toprule
            \multirow{2}[0]{*}{\textbf{Method}} & \multirow{2}[0]{*}{\textbf{Delivery(\%)}} & \multicolumn{4}{c}{\textbf{Accuracy(\%)}} \\
            \cmidrule(lr){3-6}
                  &       & \textbf{Good} & \textbf{CFactuals} & \textbf{Unreachable} & \textbf{Avg} \\
        \midrule
        \textbf{CoT} & 100.00   & 87.20    & 53.60    & 56.00    & 65.60 \\
        \midrule
        \textbf{Single Phase} & \multirow{2}[0]{*}{100.0} & \multirow{2}[0]{*}{80.00} & \multirow{2}[0]{*}{80.00} & \multirow{2}[0]{*}{76.00} & \multirow{2}[0]{*}{78.67} \\
        \textbf{(Section \ref{sec:ablation_cbs})} &       &       &       &       &  \\
        \midrule
        \textbf{w/o Code} & \multirow{2}[0]{*}{100.00} & \multirow{2}[0]{*}{64.00} & \multirow{2}[0]{*}{92.00} & \multirow{2}[0]{*}{76.00} & \multirow{2}[0]{*}{77.33} \\
        \textbf{(Section \ref{sec:ablation_cg})} &       &       &       &       &  \\
        \midrule
        \textbf{w/o Feedback} & \multirow{2}[0]{*}{94.67} & \multirow{2}[0]{*}{84.00} & \multirow{2}[0]{*}{96.00} & \multirow{2}[0]{*}{76.00} & \multirow{2}[0]{*}{85.33} \\
        \textbf{(Section \ref{sec:ablation_rf})} &       &       &       &       &  \\
        \midrule
        \textbf{BeSimulator} & 100.00  & 85.60    & 96.80    & 88.80    & 90.40 \\
        \bottomrule
    \end{tabular}%

    \caption{\label{tab:ablation}    
    The ablation experiment results on BTSIMBENCH for DeepSeek-V3. ``Single-phase'' refers to the ablation for CBS. ``w/o Code'' refers to the ablation for Code-driven Reasoning. ``w/o Feedback'' refers to the ablation for Reflective Feedback.}
\end{table}%

% D.Ablation Study
\subsection{Ablation Study}
\label{sec:ablation}

In the ablation experiments, we choose DeepSeek-V3 as the base LLM. 
Subsequently, we systematically remove each mechanism from our method, enabling us to pinpoint and comprehend the specific efficacy of each mechanism. The ablation experiment results are detailed in Table \ref{tab:ablation}.

% Effectiveness of Chain of Behavior Simulation
\subsubsection{Effectiveness of Chain of Behavior Simulation}
\label{sec:ablation_cbs}
To implement the ablation analysis on the thought mode of behavior simulation, we compare the ``consider-decide-capture-transfer'' four-phase mode of CBS with the single-phase mode, which prompts LLMs to analyze the action feasibility and state transitions in only a single phase. 
We observe that, despite using the single-phase thought mode, the average accuracy across three categories increases by 13.07\% compared to the CoT baseline. This confirms that the approach, which constructs the text-based simulation environment and performs state transitions according to the control logic of BPS, can effectively simulate BPS and enhance evaluation accuracy. 
%However, by comparing with the performance of BeSimulator and analyzing the output of LLMs, we note that due to the problem complexity, the single-phase thought mode is stuck in a lack of consideration for action feasibility and effects. 
\Nan{However, our analysis of LLM outputs reveals that the single-phase thought mode, constrained by the problem's complexity, fails to sufficiently analyze action feasibility and effects, which explains its inferior performance compared to BeSimulator.} 
This underscores the importance of CBS in enhancing action feasibility analysis and capturing state transitions, particularly those indirectly connected to actions, which improves LLMs' reliability in behavior simulation.

% Effectiveness of Code-driven Reasoning
\subsubsection{Effectiveness of Code-driven Reasoning}
\label{sec:ablation_cg}
We conduct the ablation analysis on the effects of code-driven reasoning. We remove the code-driven reasoning in CBS, transforming it into semantic reasoning.
The result indicates that three category accuracy rates decrease, particularly for the Good and Unreachable categories. We find that, in the absence of code generation and execution, LLMs consistently struggle with arithmetic calculations and comparisons. For instance, LLMs are stuck in comparing the numerical values of 1.414 and 1.0. This highlights the efficacy of the code-driven reasoning mechanism in addressing the numerical hallucination of LLMs. 
Furthermore, compared to the single-phase mode, ``w/o Code'' performs excellently in the Counterfactuals category due to the human-like thought paradigm of CBS.

% Effectiveness of Reflective Feedback
\subsubsection{Effectiveness of Reflective Feedback}
\label{sec:ablation_rf} 
The results of the ablation experiment on reflective feedback reveal a significant 5.33\% decline in delivery rate when reflective feedback is removed. 
The result shows that LLMs face challenges in providing outputs that satisfy syntax requirements and semantic consistency in one response. 
This phenomenon substantiates the significance of reflective feedback, which narrows the gap between LLMs and idea outputs and refines simulation.
\section{Conclusion}

We formalize the simulation problems for behavior planning solutions to evolve the real-world simulation challenges. 
To enhance the efficiency and generalization of simulation, we focus on behavior simulation in robotics and propose a novel LLM-based framework, BeSimulator, as an effort toward behavior simulation in the context of text-based environments. 
BeSimulator first generates text-based simulation scenes, then performs semantic-level simulation and ultimately evaluates. 
We integrate mechanisms including Chain of Behavior Simulation, code-driven reasoning, and reflective feedback to ensure the effectiveness of the simulation. 
Experimental results across four LLMs on our proposed BTs simulation benchmark BTSIMBENCH demonstrate that BeSimulator achieves significant improvements in the simulation performance for long-horizon tasks while enhancing efficiency.
\section{Limitations}

% BPS
To prove the efficacy of our work, we perform adequate experiments based on BTs.
We adopt BTs due to their popularity as a robot control architecture in recent years.
Although our results substantiate the effectiveness and efficiency of the proposed approach on BTs simulation, the performance has not been evaluated on other robot control architectures, such as FSMs, HTNs.
An important direction for future research is to extend the application of this work to a broader range of robot control architectures.

% application
In this work, we propose BeSimulator, a behavior-level simulator in the context of text-based environments, which has a different scope from the existing physics-based and visual simulation tools. 
These tools are still essential for achieving high-fidelity physical simulations, as well as in scenarios where visual information is crucial.
While the primary goal of this work is to realize efficient and versatile simulation. 
By leveraging this work, robotic system developers can conduct preliminary evaluations and optimizations prior to employing computationally intensive, resource-demanding, and costly conventional simulation tools, thereby significantly reducing development costs and expediting the overall robotics development cycle.

\section{Ethics Statement}
We recognize and ensure that our study aligns with the established Code of Ethics. 
The focus of this article is on a novel LLM-powered framework that conducts behavior simulation in text-based environments.
However, we acknowledge that as an LLM application, it may be exploited by malicious individuals. 
For example, if someone uses our work to simulate criminal methods, ethical concerns will arise.
Therefore, we urge that such applications undergo security checks on user instructions when put into use, to identify malicious attempts.

\bibliography{custom.bib}

\vfill
\newpage

\appendix
\section{PROMPT}
\label{appendix:PROMPT}

The complete prompt templates of CBS's four phases are provided in the following.

% Think of CBS
\subsection{\textit{Think} of CBS}

\begin{tcolorbox}[fonttitle=\small\fon{pbk}\bfseries,
fontupper=\scriptsize\sffamily,
fontlower=\fon{put},
enhanced,
left=2pt, right=2pt, top=2pt, bottom=2pt,
title=Think Phase Prompt]
\begin{lstlisting}[language=prompt]
You are a world model that can recognize and understand various scenes in the real world well, and can determine whether the action could be executed successfully.

### Task Description
Based on your understanding of the current world state, and semantics of the given action, your task is to determine whether this provided action could be executed successfully based on the current states and action description, and gives your reason process. 

### Input Description
I will give you 
(1) the current world state in dictionary (denoted as *Current States*). 
(2) the detailed description of current world states in dictionary (denoted as *Current States Description*) 
(3) the action description (denoted as *Action*)

### Output Rules
1. Your output must be a dictionary. Just Three keys are included: 'thought', 'corecondition', and 'corecondition_successtag'. Please do not output irrelevant content.
2. In the 'thought' key, you should first summarize the conditions that need to be met and the corresponding boolean value, based on the current world states, for the action to be executed successfully. Then, identify which states in the current world states are crucial for influencing each condition. Boolean value is true, indicating that these conditions should be met for the action execution; Boolean value is false, indicating that these conditions should not be met for the action execution. You should express it as 'condition?=true' or 'condition?=false'.
3. In the 'corecondition' key, the value is a dictionary. The dictionary includes all preconditions that affect the execution of the action. The keys of the dictionary should be expressed as complete question sentences, representing each precondition. The corresponding values should be the core states from the current world states that are necessary to check each precondition. The state names should be represented as A-B-C. Keys from different levels are connected with hyphen. Each condition corresponds to several states in a list.
4. In the 'corecondition_successtag' key, it shows the boolean value that each precondition in the 'corecondition' dictionary should return for the action to be executed successfully. Ensure that information in 'corecondition_successtag' should be consisted with the meaning in 'thought'.
5. The response should be output in the JSON format as shown in the example below, which should begins with ```json and ends with ```.

***** Example *****
{SHOTS}
***** Example Ends *****

*Current States*:
{CURRENT_STATES}
*Current States Description*:
{CURRENT_STATE_DESCRIPTION}
*Action*:
{ACTION_DESCRIPTION}
*Output*:
\end{lstlisting}
\end{tcolorbox}
\vfill
\newpage

% Decide of CBS
\titlespacing*{\subsection}{0pt}{0pt}{0pt}
\subsection{\textit{Decide} of CBS}
\vspace{-20em}
\begin{tcolorbox}[fonttitle=\small\fon{pbk}\bfseries,
fontupper=\scriptsize\sffamily,
fontlower=\fon{put},
enhanced,
left=2pt, right=2pt, top=2pt, bottom=2pt,
title=Decide Phase Prompt (Code-driven reasoning mode)]
\begin{lstlisting}[language=prompt]
You are a world model that can recognize and understand various scenes in the real world well, and can determine whether the action could be executed successfully.

### Task Description
Based on your understanding of the current world state, the semantics of the given action, and a condition related to whether the action can be executed. Your task is to determine whether the condition is true based on the current states and the provided core states, and gives your reason process. 

### Input Description
I will give you 
(1) the current world state in dictionary (denoted as *Current States*). 
(2) the detailed description of current world states in dictionary (denoted as *Current States Description*) 
(3) the action description (denoted as *Action*)
(4) one condition related to whether the action can be executed successfully (denoted as *Condition*).
(5) the core states which are key basis for you to determine whether this condition is true or false (denoted as *Core States*)

### Output Rules
1. Your output must be a dictionary. Just Two keys are included: "thought" and "code". Please do not output irrelevant content.
2. In the 'thought' key, you should give your reasoning process to make the decision based on the current world states and the core states. 
3. In the 'code' key, you must generate python codes to calculate, according to these values of core states. Specifically, the codes should be between with '###python' and ends with '###'. And lastly must get a boolean variable "resp" in final. 
4. The response should be output in the JSON format as shown in the example below, which should begins with ```json and ends with ```. 

***** Example *****
{SHOTS}
***** Example Ends *****

*Current States*:
{CURRENT_STATES}
*Current States Description*:
{CURRENT_STATE_DESCRIPTION}
*Action*:
{ACTION_DESCRIPTION}
*Condition*:
{PRECONDITION}
*Core States*:
{CORESTATES}
*Output*:
\end{lstlisting}
\end{tcolorbox}

\pagebreak

% Capture of CBS
\subsection{\textit{Capture} of CBS}
\setlength {\parskip} {0.2cm}
\begin{tcolorbox}[fonttitle=\small\fon{pbk}\bfseries,
fontupper=\scriptsize\sffamily,
fontlower=\fon{put},
enhanced,
left=2pt, right=2pt, top=2pt, bottom=2pt,
title=Capture Phase Prompt]
\begin{lstlisting}[language=prompt]
You are a world model that can recognize and understand various scenes in the real world well, and can predict future situations.

### Task Description
Based on your understanding of the current world state, and semantics of the given action, your task is to output all states needed to be changed after the action is executed, and gives your reason process. 
You need to consider the attributes of agent, objects, world environment, as well as the complex intersections of their relationships.

### Input Description
I will give you 
(1) the current world state in dictionary (denoted as *Current States*). 
(2) the detailed description of current world states in dictionary (denoted as *Current States Description*) 
(3) the action description (denoted as *Action*)

### Output Rules
1. Your output must be a dictionary. Just Two keys are included: "states_transfer" and "thought". Please do not output irrelevant content.
2. For "thought" key, please output your thought and reason process about which states needed to be changed.
3. For "states_transfer" key, its value is a list of all states from the current world states that are changed by this action, with considering of the world common knowledge and the affliation relationship provided in the current states. The state name should be represented as A-B-C. Keys from different levels are connected with short lines. Considering the dependency relationship between the states needed to be changed, these states should be output sequentially in the order in which they are updated. If there are no states needed to be changed, please output ['None'].
4. The response should be output in the standard JSON format as shown in the example below.

***** Example *****
{SHOTS}
***** Example Ends *****

*Current States*:
{CURRENT_STATES}
*Current States Description*:
{CURRENT_STATE_DESCRIPTION}
*Action*:
{ACTION_DESCRIPTION}
*Output*:
\end{lstlisting}
\end{tcolorbox}

\setlength {\parskip} {14cm}
% Transfer of CBS
\subsection{\textit{Transfer} of CBS}
\setlength {\parskip} {0.2cm}
Note that the *Thought* part of the following prompt is from the output of the capture phase.
\begin{tcolorbox}[fonttitle=\small\fon{pbk}\bfseries,
fontupper=\scriptsize\sffamily,
fontlower=\fon{put},
enhanced,
left=2pt, right=2pt, top=2pt, bottom=2pt,
title=Transfer Phase Prompt (Semantic reasoning mode)]
\begin{lstlisting}[language=prompt]
You are a world model that can recognize and understand various scenes in the real world well, and can predict future situations.

### Task Description
Based on your understanding of the current world state, and semantics of the given action, your task is to simulate the execution of the given action and predict the transition of the designated world state after the action happens. You need to consider the attributes of agent, objects, world environment, as well as the complex intersections of their relationships. Consider the immediate and potential future consequences of each action iteratively, which must be realistic and meet real-world physical laws.

### Input Description
I will give you 
(1) the current world state (denoted as *Current States*) 
(2) the detailed description of the current world state (denoted as *Current States Description*)
(3) the action description (denoted as *Action*)
(4) the complete thought about the action and its effects on the current states (denoted as *Thought*)
(5) the state you need to change after this action is executed (denoted as *States To Be Transferred*)

### Output Rules
1. You need to output the new state of '{STATE_TO_TRANSFER}' based on the information in *Thought*.
2. The response should be output in the standard JSON format as shown in the example below.

***** Example *****
{SHOTS}
***** Example Ends *****

*Current States*:
{CURRENT_STATES}
*Current States Description*:
{CURRENT_STATE_DESCRIPTION}
*Action*:
{ACTION_DESCRIPTION}
*Thought*:
{THOUGHT}
*States to be Transferred*:
{STATE_TO_TRANSFER}
*Output*:

\end{lstlisting}
\end{tcolorbox}
% Appendix B
\section{BTSIMBENCH}
\label{appendix:BTSIMBENCH}
We select two activity examples from the benchmark to demonstrate its long-horizon nature and diverse behavior types, as shown in Table \ref{tab:cleanpool} and \ref{brewcoffee}.

% Appendix C
\section{Detailed Quantitative Results}
\label{appendix:Detailed_Exp_Results}

This section presents a detailed quantitative analysis of our framework, evaluating its performance on two key dimensions: (1) the accuracy of extracting action preconditions and 
(2) the accuracy of updating world states after action execution. 
The results are summarized in Table \ref{tab:add_evaluation}.

The results reveal a consistent trend across four LLMs: the framework demonstrates substantially higher performance in updating world states than in extracting action preconditions. 
Through in-depth analysis, we find that LLMs are prone to hallucinations about real-world physical rules and generate physically implausible statements when reasoning about preconditions.
This issue manifests in three categories: missing preconditions, redundant preconditions and incorrect preconditions.
For example, for the action "\textit{open\_box}", one of the preconditions should be "\textit{is\_box\_inside\_robot\_gripper\_contact?=true}", but the LLM generates the precondition "\textit{is\_box\_inside\_robot\_gripper\_contact?=false}", which does not conform to the physical rules.
In the context of world state updating, we identify two main types of errors: missing state transitions and incorrect state transitions.
A representative failure case is shown in Appendix \ref{appendix:CASE_STUDY}.

\begin{table}[tbp]
    \centering
    \small
    \setlength{\tabcolsep}{4pt}

    \begin{tabular*}{0.43\textwidth}{l c cccc}
    
        \toprule
            Model                   & Acc\_AP(\%)          & Acc\_WS(\%)     \\
        \midrule
            Claude-3.5-Sonnet       & 94.67               & 98.67          \\
        \midrule
            DeepSeek-V3             & 96.00               & 97.33          \\
        \midrule
            Qwen2-72B-Instruct      & 78.67               & 94.67          \\
        \midrule
            Llama3.1-70B-Instruct   & 85.33               & 88.00          \\
        \bottomrule
        
    \end{tabular*}

    \caption{\label{tab:add_evaluation}    
    The quantitative results in two dimensions. ``Acc\_AP'' refers to the accuracy of extracting action preconditions. ``Acc\_WS'' refers to the accuracy of updating world states.}
\end{table}

% Appendix D
\section{FAILURE CASE STUDY}
\label{appendix:CASE_STUDY}
Due to the inherent biases and hallucinations of LLMs, the simulation may occasionally exhibit unreliability, resulting in failure cases.
Specifically, the failures primarily stem from LLMs considering redundant/irrelevant action preconditions and incorrect state transitions.
This highlights the need for further improvement in the LLM's understanding of real-world physical commonsense. 
We select one representative case for detailed demonstration, where the DeepSeek-V3 misclassifies a "Good" BT as the "Counterfactuals" category, as illustrated in Figure \ref{fig:getdrink} and Table \ref{tab:getdrink}.

\onecolumn

    \centering
    \small
    \setlength{\tabcolsep}{2.4mm}
    
    \begin{longtable}{m{0.14\textwidth} m{0.80\textwidth}}
        
        \toprule
        Task Name                  & CleanPool         \\  

        \midrule
        Task Desciption            & The robot task is to clean a stained pool with a brush and detergent. The behavior logic of a robot should be as follows. The brush and detergent are on the floor. Robot need to use a brush and detergent to scrub the pool and then rinses the pool to make it clean. The goal is to make the pool clean. The robot has two grippers, which one gripper can hold one object at a time.                     \\

        \midrule        
        \multirow{12}{*}{Action Nodes}   &
        \textbf{Move\_to\_brush}: Robot moves to the location of the brush.\\ 
        & \textbf{Pick\_up\_brush}: Robot grasps and lifts the brush with it one gripper.\\ 
        & \textbf{Move\_to\_detergent}: Robot moves to the location of detergent.\\ 
        & \textbf{Pick\_up\_detergent}: Robot picks up the detergent with its one gripper.\\ 
        & \textbf{Move\_to\_pool}: Robot moves to the location of the pool.\\ 
        & \textbf{ApplyDetergent}: Robot applies detergent which is in its gripper to the pool surface.\\ 
        & \textbf{ScrubPoolWithBrush}: Robot extends the gripper which holds a brush to the pool which has been applied with detergent, and uses the brush to scrub the pool.\\ 
        & \textbf{Place\_brush\_detergent}: Robot releases its gripper and places the brush and detergent on the floor.\\ 
        & \textbf{RinsePool}: Robot turns on the faucet, rinses the pool which has been applied with detergent and has been scrubbed, to make the pool clean. \\ 
        
        \midrule                                                                                                
        \multirow{9}{*}{Condition Nodes}   &           
        \textbf{Brush\_in\_gripper?}: Check if the brush is in the robot's gripper.\\ 
        & \textbf{Detergent\_in\_gripper?}: Check if the detergent is in the robot's gripper.\\ 
        & \textbf{IsNearBrush?}: Check if the robot is near the brush. Require a distance less than the gripper contact range of the robot to be considered near.\\ 
        & \textbf{IsNearDetergent?}:  Check if the robot is near the detergent. Require a distance less than the gripper contact range of the robot to be considered near.\\ 
        & \textbf{IsNearPool?}:  Check if the robot is near the pool. Require a distance less than the gripper contact range of the robot to be considered near.\\ 
        & \textbf{IsfaucetOpen?}: Check if the faucet is open.    \\                                               
        
        \midrule
        \multirow{35}{*}{BT (Good)}   &
        <Sequence class="SequenceNode" instance\_name="clean\_pool\_sequence"> \\
        & \quad<Fallback class="FallbackNode" instance\_name="hold\_brush"> \\
        & \quad\quad<Condition class="Brush\_in\_gripper?" instance\_name="Brush\_in\_gripper?"></Condition> \\
        & \quad\quad<Sequence class="SequenceNode" instance\_name="hold\_brush"> \\
        & \quad\quad\quad<Fallback class="FallbackNode" instance\_name="move\_to\_brush"> \\
        & \quad\quad\quad\quad<Condition class="IsNearBrush?" instance\_name="IsNearBrush?"></Condition> \\
        & \quad\quad\quad\quad<Action class="move\_to\_brush" instance\_name="move\_to\_brush"></Action> \\
        & \quad\quad\quad</Fallback> \\
        & \quad\quad\quad<Action class="pick\_up\_brush" instance\_name="pick\_up\_brush"></Action> \\
        & \quad\quad</Sequence> \\
        & \quad</Fallback> \\
        & \quad<Fallback class="FallbackNode" instance\_name="hold\_detergent"> \\
        & \quad\quad<Condition class="Detergent\_in\_gripper?" instance\_name="Detergent\_in\_gripper?"></Condition> \\
        & \quad\quad<Sequence class="SequenceNode" instance\_name="hold\_detergent"> \\
        & \quad\quad\quad<Fallback class="FallbackNode" instance\_name="move\_to\_detergent"> \\
        & \quad\quad\quad\quad<Condition class="IsNearDetergent?" instance\_name="IsNearDetergent?"></Condition> \\
        & \quad\quad\quad\quad<Action class="move\_to\_detergent" instance\_name="move\_to\_detergent"></Action> \\
        & \quad\quad\quad</Fallback> \\
        & \quad\quad\quad<Action class="pick\_up\_detergent" instance\_name="pick\_up\_detergent"></Action> \\
        & \quad\quad</Sequence> \\
        & \quad</Fallback> \\
        & \quad<Fallback class="FallbackNode" instance\_name="move\_to\_pool"> \\
        & \quad\quad<Condition class="IsNearPool?" instance\_name="IsNearPool?"></Condition> \\
        & \quad\quad<Action class="move\_to\_pool" instance\_name="move\_to\_pool"></Action> \\
        & \quad</Fallback> \\
        & \quad<Sequence class="SequenceNode" instance\_name="clean\_pool\_sequence"> \\
        & \quad\quad<Action class="ApplyDetergent" instance\_name="ApplyDetergent"></Action> \\
        & \quad\quad<Action class="ScrubPoolWithBrush" instance\_name="ScrubPoolWithBrush"></Action> \\
        & \quad\quad<Action class="Place\_brush\_detergent" instance\_name="Place\_brush\_detergent"></Action> \\
        & \quad\quad<Fallback class="FallbackNode" instance\_name="turn\_on\_faucet"> \\
        & \quad\quad\quad<Condition class="IsfaucetOpen?" instance\_name="IsfaucetOpen?"></Condition> \\
        & \quad\quad\quad<Action class="RinsePool" instance\_name="RinsePool"></Action> \\
        & \quad\quad</Fallback> \\
        & \quad</Sequence> \\
        & </Sequence> \\
        \midrule
        \pagebreak

        \midrule
        \multirow{23}{*}{BT (Counterfactuals)}   &
        <Sequence class="SequenceNode" instance\_name="clean\_pool\_sequence"> \\
        & \quad<Fallback class="FallbackNode" instance\_name="hold\_brush"> \\
        & \quad\quad<Condition class="Brush\_in\_gripper?" instance\_name="Brush\_in\_gripper?"></Condition> \\
        & \quad\quad<Action class="pick\_up\_brush" instance\_name="pick\_up\_brush"></Action> \\
        & \quad</Fallback> \\
        & \quad<Fallback class="FallbackNode" instance\_name="hold\_detergent"> \\
        & \quad\quad<Condition class="Detergent\_in\_gripper?" instance\_name="Detergent\_in\_gripper?"></Condition> \\
        & \quad\quad<Action class="pick\_up\_detergent" instance\_name="pick\_up\_detergent"></Action> \\
        & \quad</Fallback> \\
        & \quad<Fallback class="FallbackNode" instance\_name="move\_to\_pool"> \\
        & \quad\quad<Condition class="IsNearPool?" instance\_name="IsNearPool?"></Condition> \\
        & \quad\quad<Action class="move\_to\_pool" instance\_name="move\_to\_pool"></Action> \\
        & \quad</Fallback> \\
        & \quad<Sequence class="SequenceNode" instance\_name="clean\_pool\_sequence"> \\
        & \quad\quad<Action class="ApplyDetergent" instance\_name="ApplyDetergent"></Action> \\
        & \quad\quad<Action class="ScrubPoolWithBrush" instance\_name="ScrubPoolWithBrush"></Action> \\
        & \quad\quad<Action class="Place\_brush\_detergent" instance\_name="Place\_brush\_detergent"></Action> \\
        & \quad\quad<Fallback class="FallbackNode" instance\_name="turn\_on\_faucet"> \\
        & \quad\quad\quad<Condition class="IsfaucetOpen?" instance\_name="IsfaucetOpen?"></Condition> \\
        & \quad\quad\quad<Action class="RinsePool" instance\_name="RinsePool"></Action> \\
        & \quad\quad</Fallback> \\
        & \quad</Sequence> \\
        & </Sequence> \\

        \midrule
        \multirow{31}{*}{BT (Unreachable)}   &
        <Sequence class="SequenceNode" instance\_name="clean\_pool\_sequence"> \\ 
        & \quad<Fallback class="FallbackNode" instance\_name="hold\_brush"> \\
        & \quad\quad<Condition class="Brush\_in\_gripper?" instance\_name="Brush\_in\_gripper?"></Condition> \\
        & \quad\quad<Sequence class="SequenceNode" instance\_name="hold\_brush"> \\
        & \quad\quad\quad<Fallback class="FallbackNode" instance\_name="move\_to\_brush"> \\
        & \quad\quad\quad\quad<Condition class="IsNearBrush?" instance\_name="IsNearBrush?"></Condition> \\
        & \quad\quad\quad\quad<Action class="move\_to\_brush" instance\_name="move\_to\_brush"></Action> \\
        & \quad\quad\quad</Fallback> \\
        & \quad\quad\quad<Action class="pick\_up\_brush" instance\_name="pick\_up\_brush"></Action> \\
        & \quad\quad</Sequence> \\
        & \quad</Fallback> \\
        & \quad<Fallback class="FallbackNode" instance\_name="hold\_detergent"> \\
        & \quad\quad<Condition class="Detergent\_in\_gripper?" instance\_name="Detergent\_in\_gripper?"></Condition> \\
        & \quad\quad<Sequence class="SequenceNode" instance\_name="hold\_detergent"> \\
        & \quad\quad\quad<Fallback class="FallbackNode" instance\_name="move\_to\_detergent"> \\
        & \quad\quad\quad\quad<Condition class="IsNearDetergent?" instance\_name="IsNearDetergent?"></Condition> \\
        & \quad\quad\quad\quad<Action class="move\_to\_detergent" instance\_name="move\_to\_detergent"></Action> \\
        & \quad\quad\quad</Fallback> \\
        & \quad\quad\quad<Action class="pick\_up\_detergent" instance\_name="pick\_up\_detergent"></Action> \\
        & \quad\quad</Sequence> \\
        & \quad</Fallback> \\
        & \quad<Fallback class="FallbackNode" instance\_name="move\_to\_pool"> \\
        & \quad\quad<Condition class="IsNearPool?" instance\_name="IsNearPool?"></Condition> \\
        & \quad\quad<Action class="move\_to\_pool" instance\_name="move\_to\_pool"></Action> \\
        & \quad</Fallback> \\
        & \quad<Sequence class="SequenceNode" instance\_name="clean\_pool\_sequence"> \\
        & \quad\quad<Action class="ApplyDetergent" instance\_name="ApplyDetergent"></Action> \\
        & \quad\quad<Action class="ScrubPoolWithBrush" instance\_name="ScrubPoolWithBrush"></Action> \\
        & \quad</Sequence> \\
        & </Sequence>   \\
        
        \bottomrule  
        \caption{\label{tab:cleanpool}
        Example 1 \textit{CleanPool} in BTSIMBENCH. \textbf{Fault of the Counterfactuals BT:} robot does not move near the brush and detergent before picking up them, which may be out of the robot's contact range. \textbf{Fault of the Unreachable BT:} robot does not rinse the pool in the end while the task goal is to make the pool clean.}
    \end{longtable}

\pagebreak

%benchmark-example 2
    \centering
    \small

    \begin{longtable}{m{0.14\textwidth} m{0.80\textwidth}}
        
        \toprule 
        Task Name                  & BrewCoffee         \\  

        \midrule
        Task Desciption            & The robot task is to brew coffee. The behavior logic of a robot should be as follows. The coffee beans are stored in a open jar on the countertop, and the water source is the sink in the kitchen.An clean and empty bottle is near the sink. A coffee machine and a mug are also on the countertop. The robot needs to use the coffee machine to brew the coffee using the coffee beans and water, and then pour the brewed coffee into the mug. The goal is to ensure that the coffee is brewed and contained in the mug. The robot has two grippers, which one gripper can hold one object at a time.     \\

        \midrule         
        \multirow{17}{*}{Action Nodes}   &               
        \textbf{Move\_to\_sink}: The robot moves to the location of the sink.\\
        & \textbf{Grasp\_water\_bottle}: The robot grasps and lifts the water bottle with its one gripper. \\ 
        & \textbf{Fill\_water\_bottle}: The robot turns on the sink switch, fills the bottle which is in its gripper with the flowing water until the bottle is full, and then turns off the sink switch. \\
        & \textbf{Move\_to\_beans\_jar}: The robot moves to the location of the coffee beans jar. \\
        & \textbf{Grasp\_coffee\_beans}: The robot grasps a sufficient amount of coffee beans from the jar with its one gripper. \\
        & \textbf{Move\_to\_machine}: The robot moves to the location of the coffee machine. \\
        & \textbf{Pour\_coffee\_beans}: The robot pours the coffee beans which are in its gripper into the coffee machine. \\
        & \textbf{Pour\_water\_to\_mach}: The robot pours the water from the bottle which is in its gripper into the coffee machine. \\
        & \textbf{Start\_coffee\_brewing}: The robot turns on the coffee machine power switch. \\
        & \textbf{Wait\_for\_coffee\_brew}: The robot waits for until the coffee machine gets the brewed coffee. \\
        & \textbf{Move\_to\_mug}: The robot moves to the location of the mug. \\
        & \textbf{Grasp\_mug}: The robot grasps and lifts the mug. \\
        & \textbf{Pour\_coffee\_into\_mug}: The robot pours the brewed coffee from the coffee machine into the mug.\\
        
        \midrule                                                                                                
        \multirow{16}{*}{Condition Nodes}   &            
        \textbf{is\_near\_beans\_jar?}: Check if the robot is near the coffee beans jar. Require a distance less than the gripper contact range of the robot to be considered near.\\
        & \textbf{beans\_in\_gripper?}: Check if the coffee beans is in the robot's gripper. \\
        & \textbf{is\_near\_sink?}: Check if the robot is near the sink. Require a distance less than the gripper contact range of the robot to be considered near. \\
        & \textbf{water\_in\_bottle?}: Check if the bottle is filled with water. \\
        & \textbf{is\_near\_machine?}: Check if the robot is near the coffee machine. Require a distance less than the gripper contact range of the robot to be considered near. \\
        & \textbf{beans\_in\_machine?}: Check if the coffee beans have been poured into the coffee machine. \\
        & \textbf{water\_in\_machine?}: Check if the water has been poured into the coffee machine. \\
        & \textbf{brewing\_started?}: Check if the power state of the coffee machine is on and start brewing coffee based on ingredients such as coffee beans and water. \\
        & \textbf{brewing\_completed?}: Check if the coffee brewing process of coffee machine has completed and the content of the coffee machine is brewed coffee. \\
        & \textbf{mug\_in\_gripper?}: Check if the mug is in the robot's gripper. \\
        & \textbf{coffee\_in\_mug?}: Check if the brewed coffee has been into the mug.    \\                                               
        
        \midrule
        \multirow{29}{*}{BT (Good)}   &
        <Sequence class="SequenceNode" instance\_name="Brew\_Coffee\_Complete\_Mug"> \\
        & \quad<Sequence class="SequenceNode" instance\_name="Prepare\_Coffee\_Brewing"> \\
        & \quad\quad<Sequence class="SequenceNode" instance\_name="Gather\_Coffee\_Ingredients"> \\
        & \quad\quad\quad<Fallback class="FallbackNode" instance\_name="Move\_to\_sink"> \\
        & \quad\quad\quad\quad<Condition class="is\_near\_sink?" instance\_name="is\_near\_sink?"></Condition> \\
        & \quad\quad\quad\quad<Action class="Move\_to\_sink" instance\_name="Move\_to\_sink"></Action> \\
        & \quad\quad\quad</Fallback> \\
        & \quad\quad\quad<Action class="Grasp\_water\_bottle" instance\_name="Grasp\_water\_bottle"></Action> \\
        & \quad\quad\quad<Fallback class="FallbackNode" instance\_name="Fill\_water\_bottle"> \\
        & \quad\quad\quad\quad<Condition class="water\_in\_bottle?" instance\_name="water\_in\_bottle?"></Condition> \\
        & \quad\quad\quad\quad<Action class="Fill\_water\_bottle" instance\_name="Fill\_water\_bottle"></Action> \\
        & \quad\quad\quad</Fallback> \\
        & \quad\quad\quad<Fallback class="FallbackNode" instance\_name="Move\_to\_coffee\_beans"> \\
        & \quad\quad\quad\quad<Condition class="is\_near\_beans\_jar?" instance\_name="is\_near\_beans\_jar?"></Condition> \\
        & \quad\quad\quad\quad<Action class="Move\_to\_beans\_jar" instance\_name="Move\_to\_beans\_jar"></Action> \\
        & \quad\quad\quad</Fallback> \\
        & \quad\quad\quad<Fallback class="FallbackNode" instance\_name="Grasp\_coffee\_beans"> \\
        & \quad\quad\quad\quad<Condition class="beans\_in\_gripper?" instance\_name="beans\_in\_gripper?"></Condition> \\
        & \quad\quad\quad\quad<Action class="Grasp\_coffee\_beans" instance\_name="Grasp\_coffee\_beans"></Action> \\
        & \quad\quad\quad</Fallback> \\
        & \quad\quad</Sequence> \\ 
        & \quad\quad<Sequence class="SequenceNode" instance\_name="Setup\_Coffee\_Machine"> \\
        & \quad\quad\quad<Fallback class="FallbackNode" instance\_name="Move\_to\_machine"> \\
        & \quad\quad\quad\quad<Condition class="is\_near\_machine?" instance\_name="is\_near\_machine?"></Condition> \\
        & \quad\quad\quad\quad<Action class="Move\_to\_machine" instance\_name="Move\_to\_machine"></Action> \\
        & \quad\quad\quad</Fallback> \\
        & \quad\quad\quad...(continued on next page) \\
        \midrule

        \midrule
        \multirow{31}{*}{BT (Good)}   &
        \quad\quad\quad... \\
        & \quad\quad\quad<Fallback class="FallbackNode" instance\_name="Pour\_coffee\_beans"> \\
        & \quad\quad\quad\quad<Condition class="beans\_in\_machine?" instance\_name="beans\_in\_machine?"></Condition> \\
        & \quad\quad\quad\quad<Action class="Pour\_coffee\_beans" instance\_name="Pour\_coffee\_beans"></Action> \\
        & \quad\quad\quad</Fallback> \\
        & \quad\quad\quad<Fallback class="FallbackNode" instance\_name="Pour\_water\_to\_mach"> \\
        & \quad\quad\quad\quad<Condition class="water\_in\_machine?" instance\_name="water\_in\_machine?"></Condition> \\
        & \quad\quad\quad\quad<Action class="Pour\_water\_to\_mach" instance\_name="Pour\_water\_to\_mach"></Action> \\
        & \quad\quad\quad</Fallback> \\
        & \quad\quad</Sequence> \\
        & \quad</Sequence> \\
        & \quad<Sequence class="SequenceNode" instance\_name="Brew\_Pour\_Coffee\_Mug"> \\
        & \quad\quad<Fallback class="FallbackNode" instance\_name="Start\_coffee\_brewing"> \\
        & \quad\quad\quad<Condition class="brewing\_started?" instance\_name="brewing\_started?"></Condition> \\
        & \quad\quad\quad<Action class="Start\_coffee\_brewing" instance\_name="Start\_coffee\_brewing"></Action> \\
        & \quad\quad</Fallback> \\
        & \quad\quad<Fallback class="FallbackNode" instance\_name="Wait\_for\_coffee\_brew"> \\
        & \quad\quad\quad<Condition class="brewing\_completed?" instance\_name="brewing\_completed?"></Condition> \\
        & \quad\quad\quad<Action class="Wait\_for\_coffee\_brew" instance\_name="Wait\_for\_coffee\_brew"></Action> \\
        & \quad\quad</Fallback> \\
        & \quad\quad<Fallback class="FallbackNode" instance\_name="Move\_to\_mug"> \\
        & \quad\quad\quad<Condition class="mug\_in\_gripper?" instance\_name="mug\_in\_gripper?"></Condition> \\
        & \quad\quad\quad<Sequence class="SequenceNode" instance\_name="hold\_mug"> \\
        & \quad\quad\quad\quad<Action class="Move\_to\_mug" instance\_name="Move\_to\_mug"></Action> \\
        & \quad\quad\quad\quad<Action class="Grasp\_mug" instance\_name="Grasp\_mug"></Action> \\
        & \quad\quad\quad\quad<Action class="Move\_to\_machine" instance\_name="Move\_to\_machine"></Action> \\
        & \quad\quad\quad</Sequence> \\
        & \quad\quad</Fallback> \\
        & \quad\quad<Fallback class="FallbackNode" instance\_name="Pour\_coffee\_into\_mug"> \\
        & \quad\quad\quad<Condition class="coffee\_in\_mug?" instance\_name="coffee\_in\_mug?"></Condition> \\
        & \quad\quad\quad<Action class="Pour\_coffee\_into\_mug" instance\_name="Pour\_coffee\_into\_mug"></Action> \\
        & \quad\quad</Fallback> \\
        & \quad</Sequence> \\
        & </Sequence> \\

        \midrule
        \multirow{25}{*}{BT (Counterfactuals)}   &
        <Sequence class="SequenceNode" instance\_name="Brew\_Coffee\_Complete\_Mug"> \\
        & \quad<Sequence class="SequenceNode" instance\_name="Prepare\_Coffee\_Brewing"> \\
        & \quad\quad<Sequence class="SequenceNode" instance\_name="Gather\_Coffee\_Ingredients"> \\
        & \quad\quad\quad<Fallback class="FallbackNode" instance\_name="Move\_to\_sink"> \\
        & \quad\quad\quad\quad<Condition class="is\_near\_sink?" instance\_name="is\_near\_sink?"></Condition> \\
        & \quad\quad\quad\quad<Action class="Move\_to\_sink" instance\_name="Move\_to\_sink"></Action> \\
        & \quad\quad\quad</Fallback> \\
        & \quad\quad\quad<Action class="Grasp\_water\_bottle" instance\_name="Grasp\_water\_bottle"></Action> \\
        & \quad\quad\quad<Fallback class="FallbackNode" instance\_name="Fill\_water\_bottle"> \\
        & \quad\quad\quad\quad<Condition class="water\_in\_bottle?" instance\_name="water\_in\_bottle?"></Condition> \\
        & \quad\quad\quad\quad<Action class="Fill\_water\_bottle" instance\_name="Fill\_water\_bottle"></Action> \\
        & \quad\quad\quad</Fallback> \\
        & \quad\quad\quad<Fallback class="FallbackNode" instance\_name="Move\_to\_coffee\_beans"> \\
        & \quad\quad\quad\quad<Condition class="is\_near\_beans\_jar?" instance\_name="is\_near\_beans\_jar?"></Condition> \\
        & \quad\quad\quad\quad<Action class="Move\_to\_beans\_jar" instance\_name="Move\_to\_beans\_jar"></Action> \\
        & \quad\quad\quad</Fallback> \\
        & \quad\quad</Sequence> \\ 
        & \quad\quad<Sequence class="SequenceNode" instance\_name="Setup\_Coffee\_Machine"> \\
        & \quad\quad\quad... (same to the above Good BT) \\
        & \quad\quad</Sequence> \\
        & \quad</Sequence> \\
        & \quad<Sequence class="SequenceNode" instance\_name="Brew\_Pour\_Coffee\_Mug"> \\
        & \quad\quad ... (same to the above Good BT) \\
        & \quad</Sequence> \\
        & </Sequence> \\

        \midrule
        \multirow{9}{*}{BT (Unreachable)}   &
        <Sequence class="SequenceNode" instance\_name="Brew\_Coffee\_Complete\_Mug"> \\
        & \quad<Sequence class="SequenceNode" instance\_name="Prepare\_Coffee\_Brewing"> \\
        & \quad\quad... (same to the above Good BT) \\
        & \quad</Sequence> \\
        & \quad<Sequence class="SequenceNode" instance\_name="Brew\_Pour\_Coffee\_Mug"> \\
        & \quad\quad<Fallback class="FallbackNode" instance\_name="Start\_coffee\_brewing"> \\
        & \quad\quad\quad<Condition class="brewing\_started?" instance\_name="brewing\_started?"></Condition> \\ 
        & \quad\quad\quad...(continued on next page) \\ 
        \midrule 
        \vfill \\
        
        \midrule 
        \multirow{15}{*}{BT (Unreachable)}   &
        \quad\quad\quad... \\        
        & \quad\quad\quad<Action class="Start\_coffee\_brewing" instance\_name="Start\_coffee\_brewing"></Action> \\
        & \quad\quad</Fallback> \\
        & \quad\quad<Fallback class="FallbackNode" instance\_name="Move\_to\_mug"> \\
        & \quad\quad\quad<Condition class="mug\_in\_gripper?" instance\_name="mug\_in\_gripper?"></Condition> \\
        & \quad\quad\quad<Sequence class="SequenceNode" instance\_name="hold\_mug"> \\
        & \quad\quad\quad\quad<Action class="Move\_to\_mug" instance\_name="Move\_to\_mug"></Action> \\
        & \quad\quad\quad\quad<Action class="Grasp\_mug" instance\_name="Grasp\_mug"></Action> \\
        & \quad\quad\quad\quad<Action class="Move\_to\_machine" instance\_name="Move\_to\_machine"></Action> \\
        & \quad\quad\quad</Sequence> \\
        & \quad\quad</Fallback> \\
        & \quad\quad<Fallback class="FallbackNode" instance\_name="Wait\_for\_coffee\_brew"> \\
        & \quad\quad\quad<Condition class="brewing\_completed?" instance\_name="brewing\_completed?"></Condition> \\
        & \quad\quad\quad<Action class="Wait\_for\_coffee\_brew" instance\_name="Wait\_for\_coffee\_brew"></Action> \\
        & \quad\quad</Fallback> \\
        & \quad</Sequence> \\
        & </Sequence> \\

        \bottomrule
        \caption{\label{brewcoffee}
        Example 2 \textit{BrewCoffee} in BTSIMBENCH. \textbf{Fault of the Counterfactuals BT:} robot does not get coffee beans before try to pour some into the coffee machine. \textbf{Fault of Unreachable BT:} robot does not pour the brewed coffee into the mug while the task goal is to ensure that the coffee is brewed and contained in the mug.}
    \end{longtable}

%case study fig
    \vspace{80pt}

    \begin{figure*}[htbp]
        \centering
        \includegraphics[width=\linewidth]{figure/bt/21_good_9482_842.pdf}
        \caption{The BT of the failure case.}
        \label{fig:getdrink}
    \end{figure*}

\vfill
\pagebreak

%case study tab
    \centering
    \small
    \setlength{\tabcolsep}{2.4mm}
    
    \begin{longtable}{m{0.14\textwidth} m{0.80\textwidth}}
        
        \toprule
        Task Name                  & GetDrink         \\  

        \midrule
        Task Desciption            & The robot task is to prepare a drink. The behavior logic of a robot should be as follows. The water glass and a straw are inside a cabinet. The pitcher filled with orange juice is inside an electric refrigerator. And an ice cube is inside a bowl which is also in the refrigerator. The robot needs to retrieve the water glass, pitcher, ice cube from their respective locations in the kitchen, then place them on the table. Then robot should fill the water glass with orange juice, add an ice cube to the glass, and place the straw inside the glass. The goal is to obtain a glass of orange juice with ice cube. \\

        \midrule        
        \multirow{23}{*}{Action Nodes}   &
        \textbf{Move\_to\_table}: The robot moves to the location of the table.\\ 
        & \textbf{Move\_to\_cabinet}: The robot moves to the location of the cabinet.\\ 
        & \textbf{Open\_cabinet}: The robot opens the cabinet door with one gripper.\\ 
        & \textbf{Retrieve\_water\_glass\_from\_cabinet}: The robot extends its an gripper into the cabinet, holds the water glass and takes it out of the cabinet.\\ 
        & \textbf{Retrieve\_straw\_from\_cabinet}: The robot extends its an gripper into the cabinet, holds a straw and takes it out of the cabinet.\\ 
        & \textbf{Place\_glass\_straw\_to\_table}: The robot extends the gripper that holds the water glass and the straw, places the water glass and straw on the table, and then retracts the grippers.\\ 
        & \textbf{Move\_to\_refrigerator}: The robot moves to the location of the refrigerator.\\ 
        & \textbf{Open\_refrigerator}: The robot opens the refrigerator door with one gripper.\\ 
        & \textbf{Retrieve\_pitcher\_from\_refrigerator}: The robot extends its an gripper into the refrigerator, holds the pitcher, and takes the pitcher out of the refrigerator.\\ 
        & \textbf{Retrieve\_ice\_cube\_from\_refrigerator}: The robot extends its an gripper into the refrigerator, holds the ice cube, and takes ice cube out of the refrigerator.\\ 
        & \textbf{Place\_ice\_to\_table}: The robot extends the gripper that holds ice cube, places ice cube on the table and then retracts the grippers.\\ 
        & \textbf{Pour\_orange\_juice}: The robot picks up the pitcher, and then pours orange juice in the pitcher into the water glass until the glass is full.\\ 
        & \textbf{Place\_pitcher\_on\_table}: The robot extends the gripper that holds pitcher, places the pitcher on the table and releases the gripper.\\ 
        & \textbf{Add\_ice\_cube}: The robot picks up the ice cube from the table and adds them to the water glass. \\ 
        & \textbf{Insert\_straw}: The robot picks up the straw from the table and inserts it into the water glass. \\ 
        
        \midrule                                                                                                
        \multirow{13}{*}{Condition Nodes}   &           
        \textbf{Prepare\_work\_is\_done?}: Check if the robot has placed ice cube, pitcher, water glass and straw on the table.\\ 
        & \textbf{Is\_near\_cabinet?}: Check if the robot is near the cabinet. Require a distance less than the gripper contact range of the robot to be considered near.\\ 
        & \textbf{Cabinet\_door\_is\_open?}: Check if the cabinet door is open.\\ 
        & \textbf{Water\_glass\_in\_gripper?}: Check if the water glass is in the robot's gripper.\\ 
        & \textbf{Straw\_in\_gripper?}: Check if the straw is in the robot's gripper.\\ 
        & \textbf{Is\_near\_table?}: Check if the robot is near the table. Require a distance less than the gripper contact range of the robot to be considered near.\\ 
        & \textbf{Is\_near\_refrigerator?}: Check if the robot is near the refrigerator. Require a distance less than the gripper contact range of the robot to be considered near.\\ 
        & \textbf{Refrigerator\_door\_is\_open?}: Check if the refrigerator door is opened.\\ 
        & \textbf{Pitcher\_in\_gripper?}: Check if the pitcher is in the robot's gripper.\\ 
        & \textbf{Ice\_cube\_in\_gripper?}: Check if the ice cube are in the robot's gripper.\\ 
        & \textbf{Water\_glass\_is\_filled\_with\_orange\_juice?}: Check if the water glass is filled with orange juice.\\ 
        & \textbf{Ice\_cube\_in\_water\_glass?}: Check if the water glass contains the ice cube.\\ 
        & \textbf{Straw\_in\_water\_glass?}: Check if the straw is inserted into the water glass.\\                                              
        \midrule   
        \multirow{1}{*}{BT Category}   &  Good Category \\
        
        \midrule   
        \multirow{1}{*}{Simulation Result}   &  Counterfactuals Category \\

        \midrule
        \multirow{7}{*}{Failure Analysis} &
        \textbf{Failure reason:} Incorrect state transition. \\
        & \textbf{Analysis:} After the robot retrieves the straw and glass and places them on the table, it retrieves ice cubes from the refrigerator. The next actions are Move\_to\_table and Place\_ice\_to\_table. For the Move\_to\_table action, the LLM identifies three states transitions, including robot-position, ice\_cube-position and relation-ice\_cube\_on\_table. The relation ice\_cube\_on\_table needs to transition from false to true, which is an incorrect transition and prevents the execution of the next Place\_ice\_to\_table action. Ultimately, the behavior tree is mistakenly classified under the Counterfactuals category. \\
        
        \bottomrule  
        \caption{\label{tab:getdrink}
        Failure case.}
    \end{longtable}

\end{document}